\documentclass[10pt,twocolumn,letterpaper]{article}

\usepackage{iccv}
\usepackage{times}
\usepackage{epsfig}
\usepackage{graphicx}
\usepackage{amsmath}
\usepackage{amssymb}

\usepackage{booktabs}
\usepackage[pagebackref=true,breaklinks=true,letterpaper=true,colorlinks,bookmarks=false]{hyperref}
\graphicspath{{./figures/}}
\usepackage{amsmath,amsfonts,bm}

\def\eqref#1{equation~\ref{#1}}
\def\1{\bm{1}}

\DeclareMathAlphabet{\mathsfit}{\encodingdefault}{\sfdefault}{m}{sl}
\SetMathAlphabet{\mathsfit}{bold}{\encodingdefault}{\sfdefault}{bx}{n}

\newcommand{\sigmoid}{\sigma}

\newcommand{\KL}{D_{\mathrm{KL}}}

\iccvfinalcopy % *** Uncomment this line for the final submission

\def\iccvPaperID{6758} % *** Enter the ICCV Paper ID here

\ificcvfinal\fi

\begin{document}

%%%%%%%%% TITLE
\title{Multimodal Image-to-Image Translation via Mutual Information Estimation and Maximization}

\author{
	Zhiwen Zuo\textsuperscript{\rm 1}  \and
	Lei Zhao\textsuperscript{\rm 1} \and
	Zhizhong Wang\textsuperscript{\rm 1}  \and
	Haibo Chen\textsuperscript{\rm 1}  \and
	Ailin Li\textsuperscript{\rm 1}  \and
    Qijiang Xu\textsuperscript{\rm 2}  \and
	Wei Xing\textsuperscript{\rm 1}  \and
	Dongming Lu\textsuperscript{\rm 1}  \and
	\small \textsuperscript{\rm 1} Zhejiang University \textsuperscript{\rm 2} Kwai Inc. \\
	\small \{zzwcs, cszhl, endywon, feng123, liailin, wxing, ldm\}@zju.edu.cn,
	xuqijiang@kuaishou.com
}
\maketitle
\ificcvfinal\thispagestyle{empty}\fi

%%%%%%%%% ABSTRACT
\begin{abstract}
Multimodal image-to-image translation (I2IT) aims to learn a conditional distribution that explores
multiple possible images in the target domain given an input image in the source domain.
%This problem is inherently ill-posed as there are usually only one or even no corresponding images
%available in the target domain for an input image in the source domain during training.
Conditional generative
adversarial networks (cGANs) are often adopted for modeling such a conditional distribution. However,
cGANs are prone to ignore the latent code and learn a unimodal distribution in conditional image
synthesis, which is also known as the mode collapse issue of GANs. To solve the problem, we
propose a simple yet effective method that explicitly estimates and maximizes
the mutual information between the latent code and the output image in cGANs by using a deep
mutual information neural estimator in this paper. Maximizing the mutual information strengthens the
statistical dependency between the latent code and the output image, which prevents the generator
from ignoring the latent code and encourages cGANs to fully utilize the latent code for synthesizing
diverse results. Our method not only provides a new perspective from information theory to improve
diversity for I2IT but also achieves disentanglement between the source domain content and the target
domain style for free.
%We also point out the connection between our method and existing works that can
%achieve multimodal I2IT and discuss the wide application domains beyond I2IT that may
%benefit from our method.
Extensive experiments under both paired and unpaired I2IT settings
demonstrate the effectiveness of our method to achieve diverse results without loss of quality. Our
code will be made publicly available soon.
\end{abstract}

%%%%%%%%% BODY TEXT
\section{Introduction}

In recent years, Generative Adversarial Networks (GANs)~\cite{goodfellow2014generative} have
emerged as a promising generative model that can capture complex and high-dimensional image data
distributions. Extended on GANs, conditional GANs (cGANs)~\cite{mirza2014conditional} which take
extra contextual information as input  are widely applied in conditional image synthesis tasks and
achieve great success, such as image to image translation~\cite{isola2017image}, super
resolution~\cite{ledig2017photo}, image inpainting~\cite{pathak2016context}, and
text-to-image synthesis~\cite{zhang2017stackgan}.

Domain mapping or image-to-image translation (I2IT) aims to learn the mapping
from
the source
image domain $\mathcal{A}$ to the  target  image domain $\mathcal{B}$. Many conditional image
synthesis tasks can be seen as special
cases of I2IT, \eg, super resolution~\cite{ledig2017photo},
colorization~\cite{larsson2016learning}, image
inpainting~\cite{pathak2016context}, and style transfer \cite{gatys2016image}.
However, many previous works \cite{isola2017image,wang2018high, liu2017unsupervised,
	zhu2017unpaired}
on I2IT only learn a deterministic mapping function.
I2IT should be capable of producing multiple possible outputs even for
a single input image, \eg, a Yosemite winter photo	may correspond	to multiple summer photos
that vary in light, the amount of clouds, or the luxuriance of vegetation.  A straightforward way to
produce diverse results for cGANs is to distill such variations	in latent noise $Z$ that can be sampled
from a simple distribution, such as an isotropic Gaussian.	However, this problem is inherently ill-posed
as there are usually only one or even no corresponding images available in the target domain for an
input image in the source domain during training. And the signal from high-dimensional and structured
input image is usually stronger than that of the low-dimensional latent noise in cGANs. It has been
reported in the literature of conditional image synthesis \cite{isola2017image, mao2019mode,
mathieu2015deep,zhu2017toward} that cGANs are prone to overlook the latent noise, which is also
known as the mode collapse issue~\cite{goodfellow2016nips, goodfellow2014generative,
salimans2016improved} of GANs.

\begin{figure}[t!]
	\centering
	\includegraphics[width=0.5\textwidth]{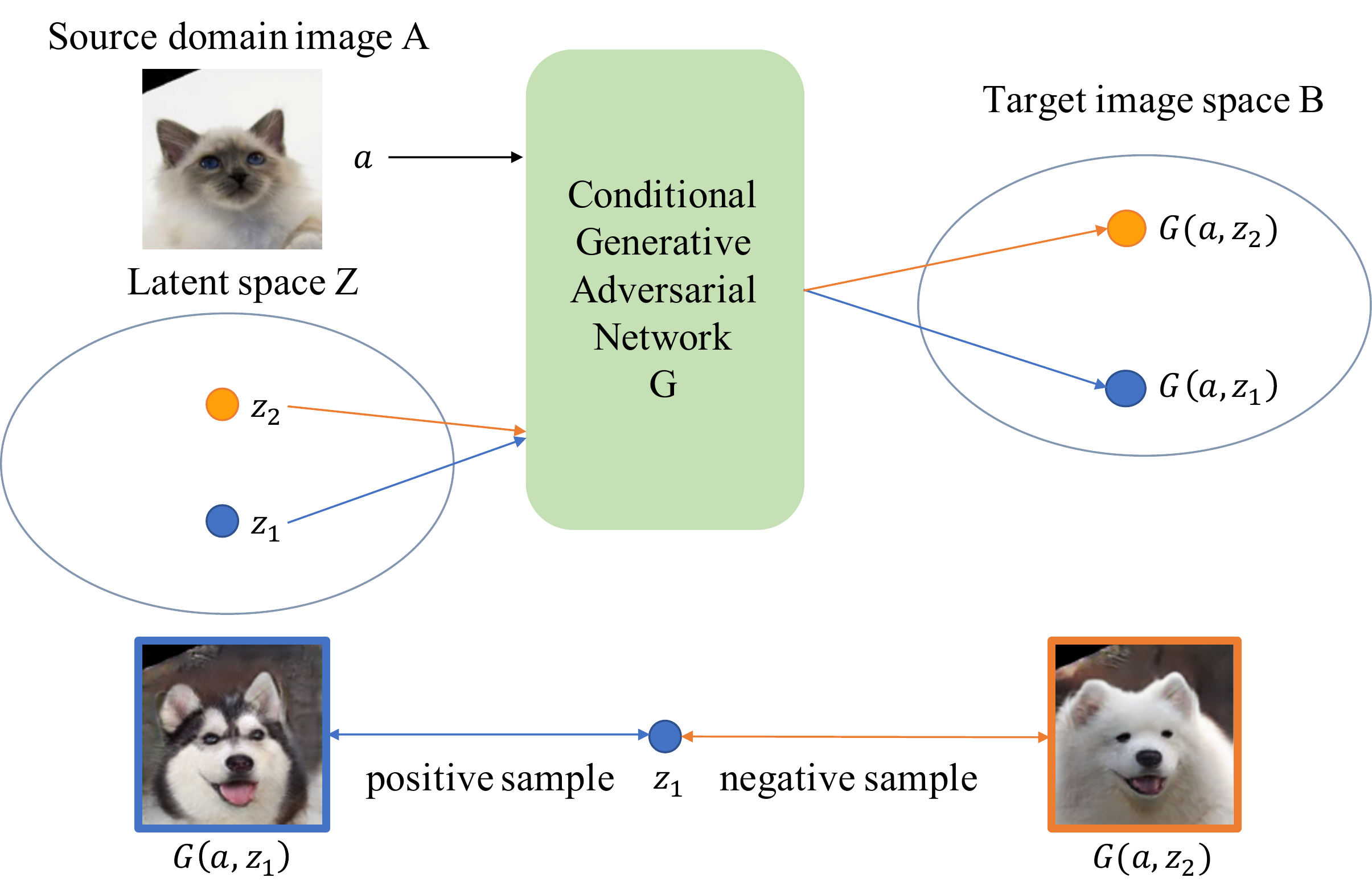}
	\caption{\textbf{Proposed method.} Our method encourages diversity for image-to-image
		translation by estimating and maximizing the deep neural mutual information between the latent
		code and the generated image in cGANs. Specifically, we classify the latent code and the images
		generated by it (\eg, $\left(z_1,G\left(a, z_1\right)\right)$) and the latent code and the images not
		generated by it (\eg, $\left(z_1,G\left(a,	z_2\right)\right)$) as the positive and negative samples
		respectively.}
	\label{fig:method}
\end{figure}

To encourage diversity for I2IT, many existing
works~\cite{almahairi2018augmented, huang2018multimodal, lee2018diverse, zhao2020uctgan,
	zhu2017toward} propose to learn a \textit{one-to-one mapping} between the latent space and the
generated image space by using informative encoders to recover the latent code from the generated
image, which we refer to as the \textit{latent code reconstruction loss} in this paper. From a
perspective of information theory, we show that this loss is highly related to the variational mutual
information maximization~\cite{barber2003algorithm,chen2016infogan} between the latent code $Z$
and the output image $\hat{B}$ as follows (refer to
supplementary materials for detailed derivation):
\begin{equation}
\mathcal{I}(Z;\hat{B}) =  H(Z) - H(Z|\hat{B}) \ge H(Z) - \mathcal{L}_R
\label{eq:MI_lower_bound}
\end{equation}
where $\mathcal{I}(\cdot;\cdot)$, $H(\cdot)$, and $\mathcal{L}_R$ are the mutual information,
the Shannon entropy and the latent code reconstruction loss, respectively. As the entropy $H(Z)$ is a
constant (the prior distribution $p_z$ is often predefined), minimizing the latent code reconstruction
loss $\mathcal{L}_R$ amounts to maximizing a variational
lower bound on the mutual information between $Z$ and $\hat{B}$. Maximizing the mutual information
improves diversity for cGANs as it enhances the statistical dependency  between $Z$ and
$\hat{B}$ and prevents the generator from ignoring $Z$.
%Therefore, the success of these works~\cite{almahairi2018augmented,
%huang2018multimodal, lee2018diverse, zhao2020uctgan, zhu2017toward} to produce diverse
%results
%for I2IT can in some degree be attributed to the variational mutual information maximization.
However, the latent code reconstruction loss is limited by the design of the specific
task~\cite{huang2018multimodal, zhao2020uctgan, zhu2017toward} and the
capacity of the encoders to learn useful information~\cite{isola2017image, larsen2016autoencoding}.
To bypass such restrictions and fully encourage the statistical dependency between the latent code
$Z$ and the output image $\hat{B}$, in this paper, we propose an alternative and straightforward way -
discarding the encoders and directly maximizing the mutual information between them. This is achieved
by the deep mutual information neural estimator~\cite{belghazi2018mine}.
Specifically, our method  introduces a statistics network $T$ to estimate and maximize the mutual
information between the latent code $Z$ and  the output image $\hat{B}=G(\cdot, Z)$ by discriminating
the corresponding positive samples $(z_1, G(\cdot, z_1))$ from the non-corresponding negative
samples $(z_1, G(\cdot, z_2))$, as shown in Figure \ref{fig:method}. Intuitively, this implies the
statistics network tries to discover the unique link between the latent code and the images generated
by it, which also indirectly decouples the target domain style from the source domain content.

We augment existing paired or unpaired I2IT methods with our proposed additional loss term that
maximizes the mutual information between the latent code and the generated image to encourage
diversity for them, \eg, pix2pix~\cite{isola2017image} and GcGAN~\cite{fu2019geometry}. A simple
architectural enhancement is shown to suffice for the utilization of our proposed loss, which results in
an elegant new model we name as Statistics Enhanced GAN (SEGAN) in this paper (see Figure
\ref{fig:model}). Both qualitative and quantitative evaluations under paired and unpaired I2IT settings
verify the effectiveness of our method for improving diversity without loss of image quality.

Our contributions in this paper are summarized below:
\begin{itemize}
	\item We propose a novel method from information theory that explicitly estimates and maximizes the
	mutual information between the latent code and the output image to palliate mode collapse in cGANs.
	\item Our method also achieves disentanglement between the source domain content and the
target domain style for free.
	\item The proposed method can facilitate many existing I2IT methods to improve diversity with a
	simple network extension.
%	Specifically, our method can be combined with one-sided unpaired I2IT
%	methods to encourage diversity for them, \eg, GcGAN, which results in a novel lightweight and
%	efficient multimodal unpaired I2IT method.
	\item Extensive experiments show the effectiveness of our method to improve diversity without
	sacrificing image quality of generated samples.
\end{itemize}

\section{Related Works}
\subsection{Generative adversarial networks}

Generative Adversarial Networks (GANs) \cite{goodfellow2014generative} compose of two
modules:
a discriminator that tries to distinguish real data samples from generated samples, and a generator
that
tries to generate samples to fool the discriminator. Many important works are proposed to improve the
original GAN for more stabilized training and producing high-quality samples, such as by proposing
better loss
functions or regularizations
\cite{arjovsky2017wasserstein,gulrajani2017improved,berthelot2017began,mao2017least,miyato2018spectral,zhao2016energy},
changing network structures
\cite{denton2015deep,karras2017progressive,karras2019style,radford2015unsupervised,zhang2019self},
or combining inference networks or autoencoders
\cite{che2016mode,donahue2016adversarial,dumoulin2016adversarially,larsen2016autoencoding,srivastava2017veegan,ulyanov2018takes}.
In this work, we rely on GANs to synthesize realistic and diverse images for multimodal image-to-image
translation.

\begin{figure*}[t!]
	\centering
	\includegraphics[width=\textwidth]{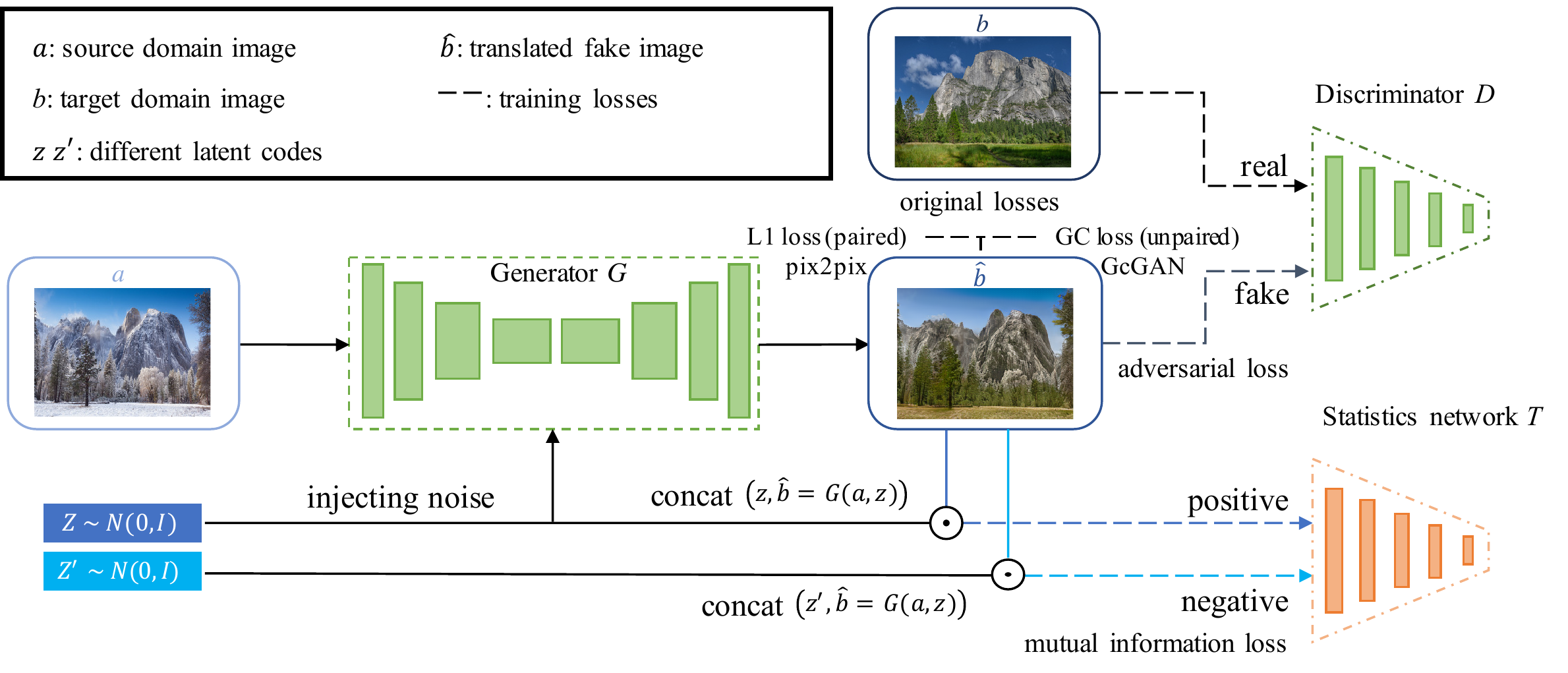}
	\caption{The network architecture and training losses of SEGAN. Our model contains two modules: a
	conditional generative adversarial network (with a generator $G$ and a discriminator $D$) and a
	statistics network $T$.  We build our method on pix2pix~\cite{isola2017image} and
	GcGAN~\cite{fu2019geometry}  under paired and unpaired settings respectively. The mutual
	information between the latent code and the output image is maximized by discriminating the
	corresponding positive samples from the non-corresponding negative samples.}
	\label{fig:model}
\end{figure*}

\subsection{Multimodal image-to-image translation}
We classify the current multimodal image-to-image translation (I2IT) methods into three categories
according to the techniques encouraging diversity for them: (1) latent code reconstruction loss,
including AugCycleGAN~\cite{almahairi2018augmented}, BicycleGAN~\cite{zhu2017toward},
MUNIT~\cite{huang2018multimodal}, DRIT~\cite{lee2018diverse}, SDIT~\cite{wang2019sdit}, and
DMIT~\cite{yu2019multi}, (2) simply maximizing the pixel difference between the generated
images, including MSGAN~\cite{mao2019mode} and
DSGAN~\cite{yang2019diversity}, and (3) the combination of both the above two techniques,
including
StarGANv2~\cite{choi2020stargan} and DRIT++~\cite{lee2020drit++}. Specifically,
SDIT~\cite{wang2019sdit}, DMIT~\cite{yu2019multi}, DRIT++~\cite{lee2020drit++}, and
StarGANv2~\cite{choi2020stargan} combine the multimodal and the multidomain I2IT in a unified
model by utilizing a domain label. However, these methods	still rely on the techniques mentioned
above to encourage diverse results. In later experiments, we will compare our method with
BicycleGAN~\cite{zhu2017toward} and MSGAN~\cite{mao2019mode} under a paired setting, and
MUNIT~\cite{huang2018multimodal}, DRIT~\cite{lee2018diverse}, and
DRIT++~\cite{lee2020drit++} under an unpaired setting, to demonstrate the effectiveness and
superiority of our proposed method.

\subsection{Mutual information}
Mutual Information (MI) is a fundamental quantity for measuring the relationship between random
variables. In contrast to correlation, mutual information can capture non-linear statistical
dependencies between variables, and thus can act as a measure of true
dependence~\cite{kinney2014equitability}.
Methods based on Mutual Information (MI) can date back to the infomax
principle~\cite{bell1995information, linsker1988self}, which advocates maximizing MI between the
input and the output of the neural networks. Notably, InfoGAN~\cite{chen2016infogan} proposed to
maximize the MI between a small part of the latent code and the generated outputs in GANs for
discovering disentangled latent representations. Our method, though shares a similar idea with
InfoGAN about maximizing the MI, yet differs from InfoGAN in two main aspects.
First, in methodology, InfoGAN minimizes a variational lower bound on the conditional entropy and
ignores the sample entropy of the MI (refer to supplementary materials), while our method
explicitly estimates and maximizes the whole MI term. Second, in target, InfoGAN aims to disentangle
latent factors in GANs, while our method mainly attacks the mode collapse issue in cGANs.
On the other hand, in BicycleGAN~\cite{zhu2017toward}, the authors attempted to use cLRGAN which
is a conditional version of InfoGAN to encourage diversity for paired I2IT, but it resulted in less variation
in the output and sometimes failed in severe mode collapse, probably because the simple latent code
reconstruction loss does not fulfill the maximization of the MI in conditional image synthesis.
In contrast, we show that our proposed method can successfully maximize the MI and therefore
improves diversity for I2IT.

Some neural estimators of MI have been proposed in recent years, such as
MINE~\cite{belghazi2018mine} and InfoNCE~\cite{oord2018representation}. Intuitively,
these methods estimate MI between random variables $X$ and $Y$ by training a classifier to
distinguish between the corresponding samples drawn from the joint distribution $p(x,y)$ and the
non-corresponding samples drawn from the product of marginals $p(x)p(y)$.
ALICE~\cite{li2017alice} revealed
that cycle consistency~\cite{kim2017learning, zhu2017unpaired, yi2017dualgan} is actually
maximizing a lower bound on the MI between the input and the output
of the generator.  Based on that, CUT~\cite{park2020contrastive} adopts the InfoNCE loss to
maximize the MI between the corresponding patches of the input and the output to preserve content
information for unpaired I2IT recently.
However, our target and methodology are totally different from CUT: (1) The
target of CUT is to better preserve the content for unpaired I2IT, which
\textit{can only produce unimodal results}. Our method aims to \textit{yield diverse results} for I2IT
and theoretically can be combined with CUT to encourage diversity for it. (2) The methodology is also
different. CUT
uses \textit{NCE loss} to maximize MI between \textit{the corresponding patches of the input and the
output images}, which cannot be used in our case. We formulate a \textit{JSD MI estimator} to
estimate and maximize the MI between \textit{the latent code and the output image}. The mathematical
form of our loss is a binary cross-entropy, which does not require a large negative sample size per
positive sample as in NCE loss; thus, it is more stable and easier to implement.

\section{Statistics Enhanced GAN}
\subsection{Preliminaries}
Suppose we have a source image domain $\mathcal{A}\subset \mathbb{R}^{\text{H}\times
	\text{W}\times 3}$ and a
target image domain $\mathcal{B}\subset \mathbb{R}^{\text{H}\times \text{W}\times 3}$,
multimodal image-to-image translation (I2IT) refers to learning a generator's function $G(a,z)$ such
that $a\in \mathcal{A}$, $\hat{b}=G(a,z)\in \mathcal{B}$ and
$\hat{b}$ preserves some underlying spatial information of $a$\footnote{We use uppercase letters
for random variables and lowercase letters for realizations of these random variables.}.  We refer to
such underlying spatial
information and the semantic difference of the target image domain as  ``content'' and ``style''
respectively.
%We further divide our framework into two aspects: 1) encouraging diverse results
%conditioned on an input image and 2) preserving the content through domain mapping. The former
%one
%is achieved by our proposed mutual inforamtion constraint.
An adversarial loss shown below is used to ensure $G(a,z)$ to look like real samples from the target
image domain. And the latent code $Z$ is responsible for the style variations of the output image
$\hat{b}$.

\begin{equation}
\begin{split}
&\max_{D_\theta} \min_{G_\gamma} \mathbb{E}_{b\sim
	p(b)}[\log
D_\theta(b)] + \\
&\mathbb{E}_{a\sim p(a),z\sim p(z)}[\log (1-D_\theta(G_\gamma(a,z)))]
\end{split}
\label{eq:GAN}
\end{equation}
where $D$, $D(\cdot)$, $\theta$, $G$, $G(\cdot,\cdot)$, and $\gamma$ are the discriminator,
the discriminator's critic function, the paramaters of the discriminator, the generator, the generator's
function, and the parameters of the generator, respectively.

For simplicity, $Z$ often follows a standard Gaussian such that $z\sim \mathcal{N}(0,I)$. After
training, we expect that varying the latent noise conditioned on an input image can produce different
images varying in style while preserving the content. However, several previous works
\cite{isola2017image, mao2019mode, mathieu2015deep, zhu2017toward}
have reported that naively adding the noise can hardly  produce diverse results, probably because of
the mode collapse issue of GANs.

\subsection{Probabilistic analysis}
To encourage diversity for image-to-image translation (I2IT), we
propose to enhance the connection between the latent noise $Z$ and the output image
$\hat{B}$ in a statistical manner by maximizing the mutual information between them, therefore we call
our proposed model Statistics Enhanced GAN (SEGAN).

\textit{Why maximizing the MI between $Z$ and $\hat{B}$ helps to reduce mode collapse in cGANs?}
Suppose the latent noise $Z$ is ignored
by the generator which is often encountered in conditional image synthesis
tasks~\cite{isola2017image, mao2019mode, mathieu2015deep, zhu2017toward}, the
generator's
function $G(a,z)$ and the conditional distribution
$p(\hat{b}|a,z)$  modeled by cGANs would degenerate to $G(a)$ and
$p(\hat{b}|a)$
respectively,
which means $Z$ is independent of $\hat{B}$ and the translation becomes
deterministic. In information theory, such dependency is measured by the Mutual Information (MI).
The MI between the latent code $Z$ and the output image $\hat{B}$
\begin{equation}
\mathcal{I}(\hat{B};Z)=H(\hat{B})-H(\hat{B}|Z)=H(Z)-H(Z|\hat{B})
\label{eq:MI}
\end{equation}
quantifies the ``amount of information'' of $\hat{B}$ through observing $Z$. When
the generator overlooks $Z$ (which means $Z$ is independent of $\hat{B}$), the MI attains its minimal
value \textbf{0}, because knowing $Z$ reveals nothing about $\hat{B}$
under
such
circumstances. In contrast, if $Z$ and $\hat{B}$ are closely related by an invertible function, \eg, $Z$ is
utilized by the generator to represent different styles of the
target image domain, then the MI attains its maximum. From another perspective, the direct
way to encourage diversity of a random variable is to maximize its entropy. As the
generated sample's entropy $H(\hat{B})$ is intractable, we use the MI between $\hat{B}$ and $Z$  as
a proxy. As shown in Equation \ref{eq:MI}, the MI can be seen as a lower bound on $H(\hat{B})$.

\subsection{Architecture and disentanglement}
SEGAN is simple and neat in its
architecture: it is merely an extension upon cGANs with a statistics network $T$ as
shown in Figure \ref{fig:model}. The statistics network is used to estimate and maximize the Mutual
Information (MI) between the latent noise $Z$ and the output image $\hat{B}$.

In our method, the statistics network $T$ estimates and maximizes the MI by
performing a
binary classification between the ``positive samples'' $\left(z, G(\cdot,
z)\right)$ (the latent code and the images generated by it)  and the ``negative samples''
$\left(z',G(\cdot,z)\right)$ (the latent code and the images not  generated by it) . This indicates the
statistics network has to find out the universal influence that the latent code $z$ puts on the output
image $G(a, z)$ regardless of the input image $a$ (because a fixed $z$ can be combined with many
different source domain images $a$). As such, the method forces the latent code to represent the style
of the target image domain instead of the noises or the structures of the content. Once successfully
trained, our method therefore achieves disentanglement between the source domain content and the
target domain style for free.

\subsection{Loss functions}

\label{sec:training_losses}

\paragraph{Mutual information loss.}
Formally, the Mutual Information (MI) between random variables $X$ and $Y$ is defined as the
Kullback-Leibler (KL)
divergence between the joint distribution $p(x,y)$ and the product of the marginals $p(x)p(y)$:

\begin{equation}
\begin{split}
\mathcal{I}(X;Y)&=\KL\left[p\left(x,y\right)\Vert
p\left(x\right)p\left(y\right)\right]
\\
&=\mathbb{E}_{p\left(x,y\right)}\left[\log
\frac{p(x,y)}{p(x)p(y)}\right]
\end{split}
\end{equation}

%\begin{equation}
%\begin{split}
%\mathcal{I}(X;Y)&:=\KL(\mathbb{J}\Vert \mathbb{M}) \ge \hat{\mathcal{I}}^{\text{DV}}_\omega(X;Y)
%\\
%&:=\mathbb{E}_\mathbb{J}\left[T_\omega(x,y)\right] - \log
%\mathbb{E}_\mathbb{M}\left[e^{T_\omega(x,y)}\right]
%\end{split}
%\label{eq:DV}
%\end{equation}
%where $T_\omega:\mathcal{X} \times \mathcal{Y} \rightarrow \mathbb{R}$ is the statistics network
%parameterized by $\omega$, $\mathbb{J}$ is the joint and
%$\mathbb{M}$ is the product of the mariginals of random variables $X$ and $Y$. This estimator
%$\hat{\mathcal{I}}^{\text{DV}}_\omega$ is trained to approximate the MI by maximizing
%Eq.\ref{eq:DV}
%w.r.t. the statistics network's parameters $\omega$.

Belghazi \etal \cite{belghazi2018mine} propose several neural estimators of MI based on the dual
formulations of the KL-divergence, that are trainable through back-propagation and highly
consistent, such as the Donsker-Varadhan (DV) representation \cite{donsker1983asymptotic} and
the f-divergence representation~\cite{nguyen2010estimating,nowozin2016f}. However, these
estimators have some defeats making them difficult to be applied for
practical use, such as biased estimate of the batch gradient and unbounded
value~\cite{belghazi2018mine}. For a more stabilized training, we adopt a neural estimator that lower
bounds the Jensen-Shannon Divergence (JSD) between the joint and the product of the marginals
following the formulation of f-GAN \cite{nowozin2016f}. This JSD MI estimator is more stable and
positively correlated with MI (refer to supplementary materials for the proof):

\begin{equation}
\begin{split}
&\text{JSD}\left[p(z,\hat{b})\Vert p(z)p(\hat{b})\right] \ge \\
&\hat{\mathcal{I}}^{\text{JSD}}_\omega(Z;G_\gamma(\cdot, Z))
=\mathbb{E}_{p_z}\left[\log
\left(\sigmoid \left(T_\omega\left(z,G_\gamma\left(\cdot,z\right)\right)\right)\right)\right]
+\\ &\mathbb{E}_{\hat{p}_z\times p_z}\left[\log \left( 1 - \sigmoid \left(T_\omega\left(z',
G_\gamma\left(\cdot,z\right)\right)\right)\right)\right]
\end{split}
\label{eq:JSD_MI}
\end{equation}
where $\hat{b}=G_\gamma(\cdot,z)$, $T_\omega(\cdot,\cdot)$ is the statistics network's function
with parameters
$\omega$ that outputs a scalar, $p_z = \hat{p}_z =\mathcal{N}(0,I)$, $z$ and $z'$  are
different samples
from $\mathcal{N}(0,I)$, and $\sigmoid(\cdot)$ is
the sigmoid function. The mathematical form of the JSD MI estimator amounts to the familiar
binary cross-entropy. The MI is estimated by maximizing $\hat{\mathcal{I}}^{\text{JSD}}_\omega$ in
Equation
\ref{eq:JSD_MI} \wrt the statistics
network's parameters $\omega$.

As maximizing MI follows the same direction of estimating MI, we jointly optimize
the generator and the statistics network by the mutual information loss below:
\begin{equation}
\mathcal{L}_{\text{MI}}= \max_{T_\omega, G_\gamma}
\hat{\mathcal{I}}_\omega^{\text{JSD}}
\left(Z;G_\gamma\left(\cdot,Z\right)\right)
\label{eq:MI_loss}
\end{equation}

\paragraph{Total loss.}
Our method can be combined with any existing I2IT methods to improve diversity for them as long as
they are built on cGANs. Thus, the final loss function for training can be formulated by adding our
proposed mutual information loss to the original losses of the existing method:
\begin{equation}
\mathcal{L}_{\text{tota}l} = \mathcal{L}_{\text{ori}} + \lambda_{\text{MI}}\mathcal{L}_{\text{MI}}
\label{eq:total_loss}
\end{equation}
where $\mathcal{L}_{\text{ori}}$ is the original losses and $\lambda_{\text{MI}}$ is the weight
that controls the importance of our mutual information loss.

%
%The L1 loss and the geometry
%consistency loss are defined respectively as follows:
%\begin{equation}
%\begin{split}
%\mathcal{L}_{gc} = &\mathbb{E}_{a\sim p(a), z\sim p(z)} [||G_\gamma(f(a),z) -
%f(G_\gamma(a,z))||_1 +\\& ||G_\gamma(a,z) - f^{-1}(G_\gamma(f(a),z))||_1]
%\end{split}
%\label{eq:L1_loss}
%\end{equation}
%where $f(\cdot)$ and $f^{-1}(\cdot)$ are the geometric transformation function and its inverse
%function
%respectively. In this paper, we use the $90^\circ$ clockwise rotation as the geometric transformation
%function.
%
%\paragraph{Total loss.}
%The total training loss functions of SEGAN under multimodal supervised or unsupervised I2IT settings
%are shown respectively as follows:
%\begin{equation}
%\mathcal{L}_{\text{paired}} = \mathcal{L}_{\text{GAN}} + \lambda_1 \mathcal{L}_{\text{MI}} +
%\lambda_2
%\mathcal{L}_{\text{L1}}
%\end{equation}
%
%\begin{equation}
%\mathcal{L}_{\text{unpaired}} = \mathcal{L}_{\text{GAN}} + \lambda_1 \mathcal{L}_{\text{MI}} +
%\lambda_3
%\mathcal{L}_{gc}
%\end{equation}
%where $\lambda_1$, $\lambda_2$ and $\lambda_3$ are the weights controlling the importance of the
%corresponding losses.

\begin{figure}[t!]
	\centering
	\includegraphics[width=\columnwidth]{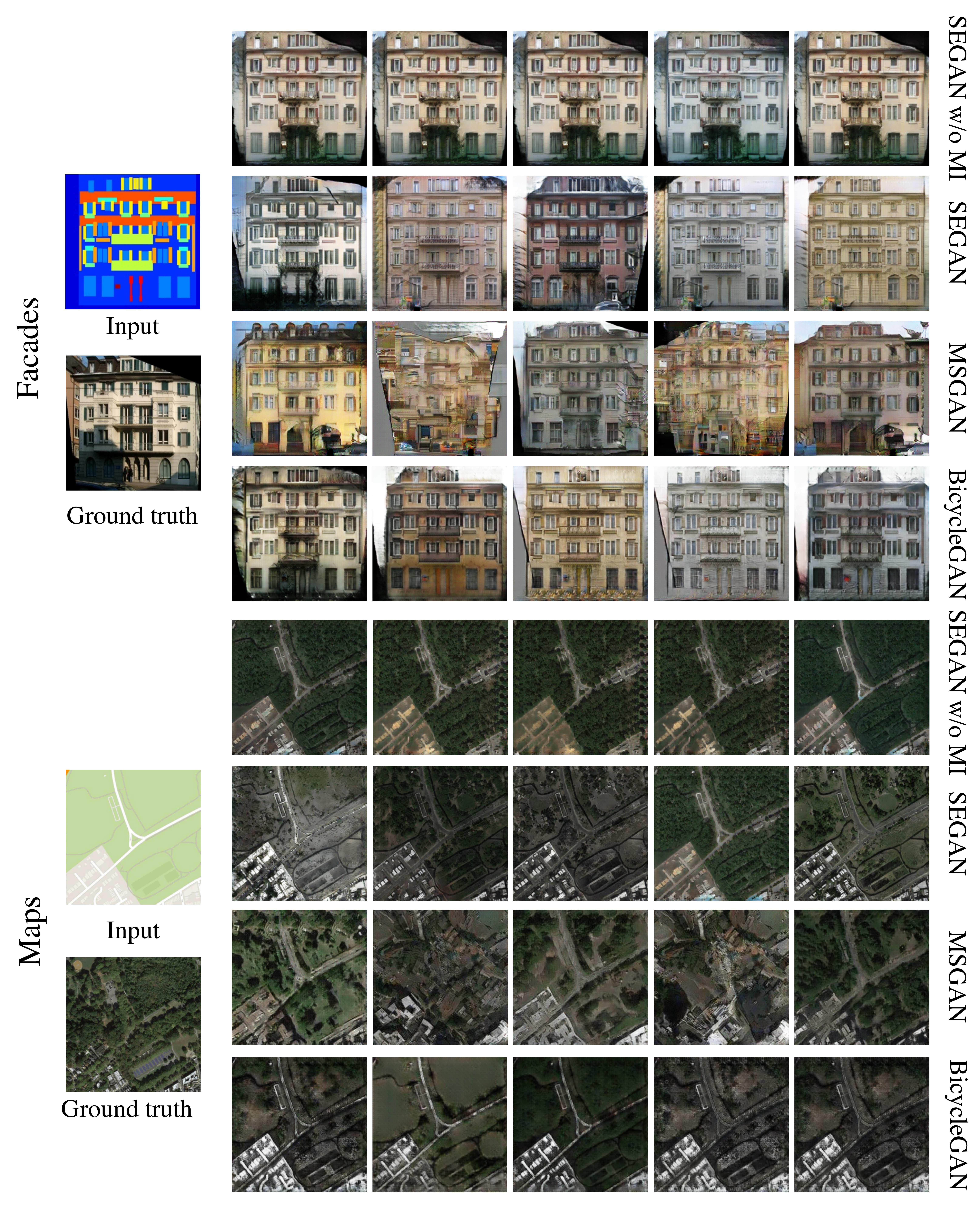}
	\caption{Qualitative results of different methods on the Facades and Maps datasets under a paired
	setting.}
	\label{fig:paired}
	\vspace{-3mm}
\end{figure}

\begin{figure*}[t!]
	\centering
	\includegraphics[width=\textwidth]{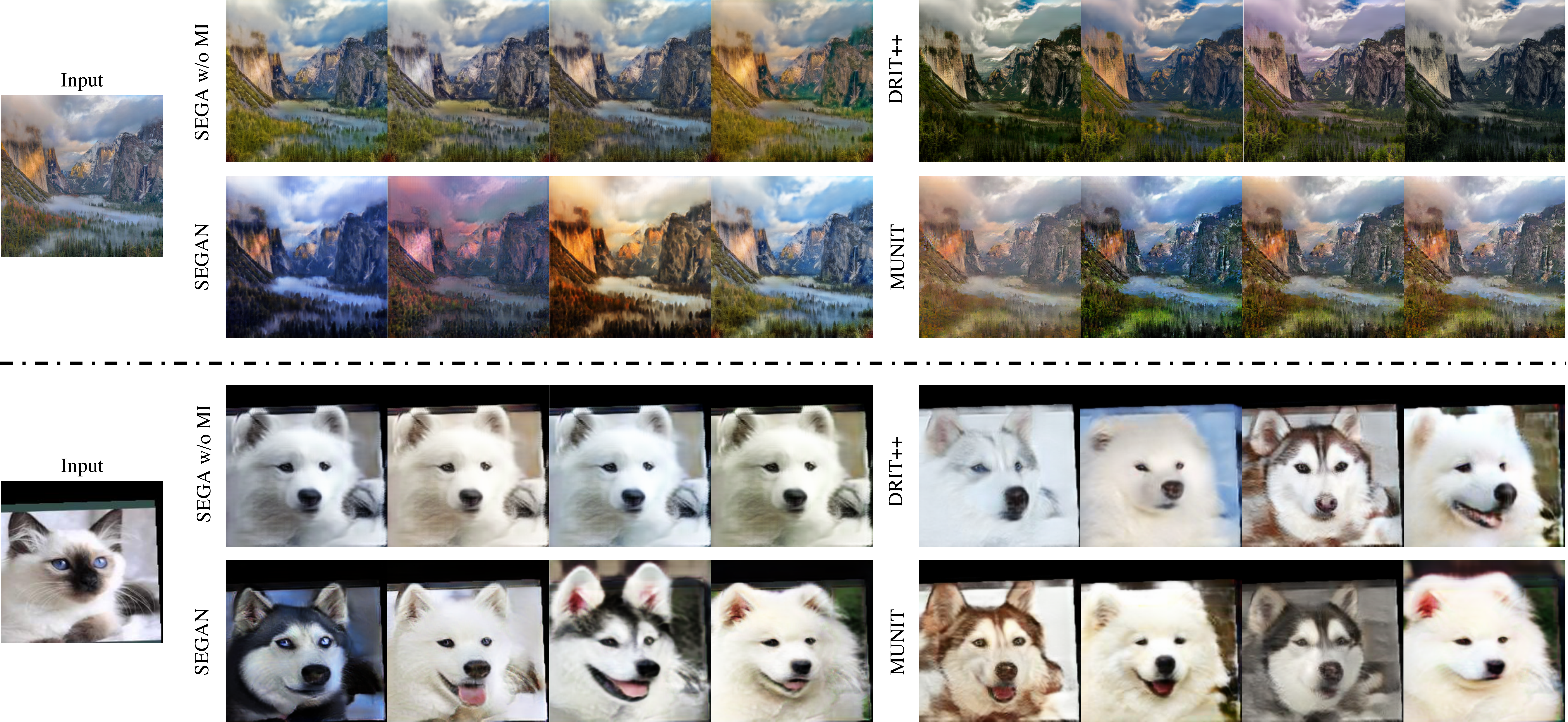}
	\caption{Qualitative results of different methods on the Yosemite winter$\rightarrow$summer and
		cat$\rightarrow$dog datasets under an unpaired setting.}
	\label{fig:unpaired}
\end{figure*}

\section{Experiments}
\subsection{Baselines and compared methods}
We evaluate our method by building it on pix2pix~\cite{isola2017image} and
GcGAN~\cite{fu2019geometry} under paired and unpaired settings respectively. Therefore, the
baselines are naively adding noise to pix2pix and GcGAN and we call them as SEGAN w/o MI. Under
a paired setting, we also compare our method (SEGAN) with two of the state-of-the-art methods:
BicycleGAN~\cite{zhu2017toward} and MSGAN~\cite{mao2019mode}. Under an unpaired setting,
we compare ours with some state-of-the-art multimodal unsupervised I2IT methods, \eg.,
MUNIT~\cite{huang2018multimodal}, DRIT~\cite{lee2018diverse}, and
DRIT++~\cite{lee2020drit++}.
\subsection{Datasets}
Under a paired setting, we evaluate on
Google maps$\rightarrow$satellite (Maps)~\cite{isola2017image} and labels$\rightarrow$images
(Facades) \cite{cordts2016cityscapes}. Under an unpaired setting, we perform experiments on a
shape-invariant dataset: Yosemite winter$\rightarrow$summer~\cite{zhu2017unpaired} and a
shape-variant dataset: cat$\rightarrow$dog~\cite{lee2018diverse}.
\subsection{Metrics}
For quantitative evaluation, we mainly use four metrics: FID~\cite{heusel2017gans},
NDB~\cite{richardson2018gans}, JSD~\cite{richardson2018gans}, and
LPIPS~\cite{zhang2018unreasonable}.

\paragraph{FID.}
FID measures the quality of generated images by calculating the distance between the
model distribution and the real one through deep features extracted by Inception
network~\cite{szegedy2015going}. Lower FID values indicate higher quality.

\paragraph{NDB and JSD.}
NDB and JSD~\cite{richardson2018gans} are two bin-based metrics. The training data is first
clustered by K-means into different bins which can be viewed as modes of the real data distribution.
Then each generated sample is assigned to the bin of its nearest neighbor. Two-sample test is
performed on each bin and then the number of statistically-different bins (NDB) is reported. The
Jensen-Shannon Divergence (JSD) between the training data's bins distribution and the generated
data's bins distribution is also reported. For both two metrics, lower values indicate higher diversity.

\paragraph{LPIPS.} LPIPS~\cite{zhang2018unreasonable} measures the averaged feature distances
between generated samples conditioned on the same input image. Higher LPIPS  score indicates better
diversity among the generated images.

\begin{table*}[t]
	\centering
	\resizebox{0.77\linewidth}{!}
	{
		%%\vspace{-1mm}
		\begin{tabular}{@{}ccccc}
			\toprule
			Dataset & \multicolumn{4}{c}{Facades}
			\\  \cmidrule(lr){2-5}
			 &  SEGAN w/o MI &  SEGAN & BicycleGAN~\cite{zhu2017toward} &
			MSGAN~\cite{mao2019mode}    \\ \cmidrule(lr){2-5}
			FID $\downarrow$ & $89.14\pm{0.99}$
			&$\mathbf{78.43\pm{1.15}}$ & $98.85\pm{1.21}$
			&$92.84\pm{1.00}$ \\
			NDB$\downarrow$  & $14.40\pm{0.65}$  &  $\bf{11.60\pm{0.80}}$ &
			$13.80\pm{0.45}$                 &
			$12.40\pm{0.55}$ \\
			JSD$\downarrow$  & $0.068\pm{0.003}$               & $\bf{0.036\pm{0.005}}$ &
			$0.058\pm{0.004}$                & $0.038\pm{0.0011}$\\
			LPIPS$\uparrow$  &
			$0.1038\pm{0.0015}$                 & $\bf{0.1897\pm{0.0015}}$ & $0.1413\pm{0.0005}$              &
			$0.1894\pm{0.0011}$
			\\ \midrule
			Dataset &  \multicolumn{4}{c}{Maps}
			\\ \cmidrule(lr){2-5}
			      &      SEGAN w/o MI &  SEGAN & BicycleGAN~\cite{zhu2017toward}   &
			      MSGAN~\cite{mao2019mode}
			\\\cmidrule(lr){2-5}
			FID$\downarrow$ &   $126.73\pm{1.47}$
			&$\mathbf{119.51\pm{0.62}}$ &  $145.78\pm{3.90}$    &$152.46\pm{2.52}$ \\
			NDB$\downarrow$ & $47.40\pm{1.50}$               &  $\bf{40.72\pm{1.00}}$&
			$46.60\pm{1.34}$
			&$41.60\pm{0.55}$   \\
			JSD$\downarrow$ & $0.048\pm{0.002}$                &$\bf{0.021\pm{0.002}}$ &
			$0.023\pm{0.002}$        &$0.031\pm{0.003}$ \\
			LPIPS$\uparrow$ &$0.1121\pm{0.002}$	&$0.1972\pm{0.003}$ &  $0.1142\pm{0.001}$
			&$\mathbf{0.2049\pm{0.002}}$\\
			\bottomrule
		\end{tabular}
	}
	\vspace{2mm}
	\caption{Quantitative results of different methods on the Facades and the
		Maps datasets under a paired setting.}
	\label{tab:paired_results}
	\vspace{-2mm}
\end{table*}

\begin{table*}[t]
	\centering
	%%\vspace{-1mm}
	\begin{tabular}{@{}ccccccc}
		\toprule
		Dataset & \multicolumn{5}{c}{Winter $\rightarrow$ Summer}
		\\  \cmidrule(lr){2-6}
		&    SEGAN w/o MI & SEGAN &   MUNIT~\cite{huang2018multimodal} &
		DRIT~\cite{lee2018diverse}         &       DRIT++~\cite{lee2020drit++}
		\\ \cmidrule(lr){2-6}
		FID $\downarrow$ & $50.05\pm{0.64}$ & $\bf{40.72\pm{0.82}}$ &  $57.09\pm{0.37}$ &
		$41.34\pm{0.20}$  &
		$41.02\pm{0.24}$
		 \\
		NDB$\downarrow$  & $10.33\pm{0.85}$ & $\bf{9.15\pm{0.67}}$ & $9.53\pm{0.64}$        &
		$9.38\pm{0.74}$ &
		$9.22\pm{0.97}$ \\
		JSD$\downarrow$  & $0.302\pm{0.052}$  & $\bf{0.210\pm{0.070}}$ &   $0.293\pm{0.062}$
		&
		$0.304\pm{0.075}$&
		$0.222\pm{0.070} $ \\
		LPIPS$\uparrow$  & $0.1032 \pm{0.0007}$ & $\bf{0.1381\pm{0.0004}}$ &
		$0.1136\pm{0.0008}$        &
		$0.0965\pm{0.0004}$ &
		$0.1183\pm{0.0007}$
		\\ \midrule
		Dataset &  \multicolumn{5}{c}{Cat $\rightarrow$ Dog}
		\\ \cmidrule(lr){2-6}
		& SEGAN w/o MI  & SEGAN  & MUNIT~\cite{huang2018multimodal}   &
		DRIT~\cite{lee2018diverse}         &       DRIT++~\cite{lee2020drit++}
		\\ \cmidrule(lr){2-6}
		FID$\downarrow$ & $28.86\pm{0.35}$ & $\bf{17.20\pm{0.22}}$ &  $22.13\pm{0.71}$ &
		$24.31\pm{0.33}$  &       $17.25\pm{0.65}$\\
		NDB$\downarrow$ &   $9.70\pm{0.51}$ & $\bf{7.52\pm{1.35}}$ &    $8.21\pm{1.17}$      &
		$8.16\pm{1.60}$ &             $7.57\pm{1.25}$\\
		JSD$\downarrow$ & $0.144\pm{0.030}$ & $\bf{0.037\pm{0.023}}$ &   $0.132\pm{0.066}$       &
		$0.075\pm{0.046}$ &
		$0.041\pm{0.014}$\\
		LPIPS$\uparrow$ & $0.231\pm{0.002}$ & $\bf{0.282\pm{0.002}}$ & $0.244\pm{0.002}$    &
		$0.245\pm{0.002}$
		&
		$0.280\pm{0.002}$\\
		\bottomrule

	\end{tabular}
	\vspace{2mm}
	\caption{Quantitative results of the Yosemite winter$\rightarrow$summer and the
		cat$\rightarrow$dog datasets under an unpaired setting.}
	\label{tab:unpaired_results}
	\vspace{-3mm}
\end{table*}

\subsection{Implementation details}
We adopt the original non-saturation GAN loss~\cite{goodfellow2014generative} for
training. As our proposed neural estimator needs a relatively large batch size to stabilize the training and
reduce variance in estimation, we use a batch size of 32 in our experiments.
%Accidently, we
%find that this relatively larger batch size for adversarial training can greatly improve the baseline
%models'
%quality and slightly increase diversity for I2IT, which is coincident with the findings of BigGAN~\cite{}.
%This breaks the common sense in the conditional image synthesis community that naively adding the
%noise to I2IT models can not produce diverse results. And surprisingly,  we observe that NSGAN loss
%can usually obtain higher diversity than LSGAN despite NSGAN is less stable than LSGAN. Although
%the
%baseline models obtain some performance gains due to the relatively increased batch size, we show
%that our method can significantly improve diversity upon them in our experiments.
After a hyperparameter search on  $\lambda_{\text{MI}}$ in preliminary experiments, we find setting it
as 3 is the best and consistently set $\lambda_{\text{MI}}=3$ in all our experiments. The statistics
network is a Convolutional Neural Network (CNN) followed by a few fully-connected layers. The choice
of sampling strategy is important to make our algorithm work. As suggested in
\cite{hjelm2018learning}, we exclude the positive samples from the product of marginals. For more
details, please refer to the supplementary materials.

\subsection{Mutual information maximization}
\begin{figure}[b!]
	\vspace{-3mm}
	\centering
	\includegraphics[width=\columnwidth]{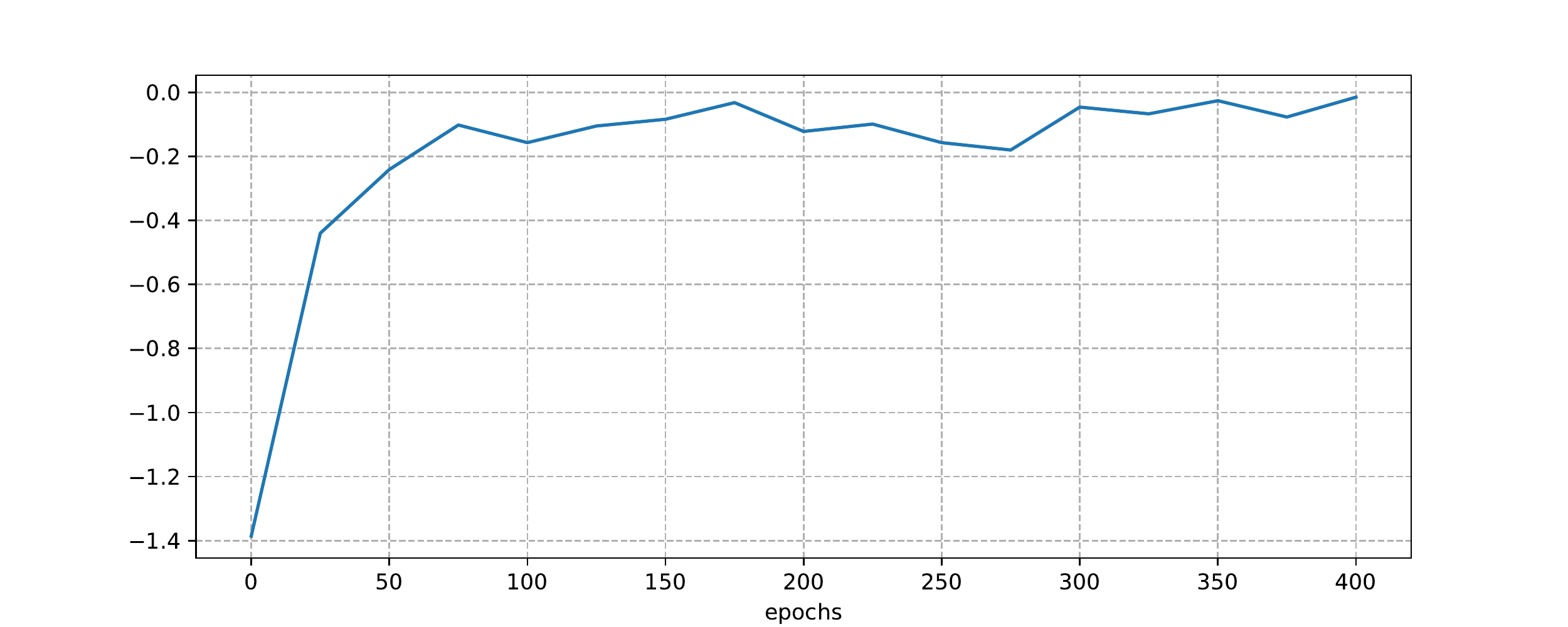}
	\caption{The change of the estimate of MI by our JSD MI estimator during training on Facades
	dataset.}
	\label{fig:MI}
	\vspace{-3mm}
\end{figure}

Figure \ref{fig:MI} shows the change of the estimate of MI by our JSD MI estimator during training
on the Facades dataset.  In the beginning, as $Z$ does not play a role in producing diverse results,
the statistics network can not tell the positive samples apart from the negative samples, and thus it
makes a random guess on them, resulting in a quantity of $\log0.5+\log0.5\approx-1.386$ (see
Equation \ref{eq:JSD_MI}). And as the training proceeds, the estimate of the MI is maximized to
$\bf{0}$, which means the statistics network can faithfully discriminate the positive samples from
negative samples.

\subsection{Paired image-to-image translation}
We build our method on pix2pix~\cite{isola2017image} under a paired I2IT setting. For a fair
comparison, all methods use the same network architecture as in BicycleGAN~\cite{zhu2017toward}.
We set the weight of L1 loss as 3 in our method. SEGAN consistently improves on all metrics over
the baseline method as shown in Table~\ref{tab:paired_results}. Note that SEGAN w/o MI has better
FIDs than BicycleGAN and MSGAN~\cite{mao2019mode}, probably due to the relatively large batch
size we used. We also tried using the same batch size for training BicycleGAN and MSGAN but obtained
no performance gains. Compared with the state-of-the-art methods, our method's diversity
outperforms BicycleGAN and is comparable to MSGAN (as measured by NDB, JSD, and LPIPS) while
with a higher quality (as measured by FID). We can also observe that MSGAN sometimes produces
images with artifacts and distortions as shown in  Figure \ref{fig:paired}, which may be caused by the
objective function of MSGAN that simply maximizes the pixel difference between output images. More
qualitative results can be found in supplementary materials.

\begin{figure*}[t!]
	\centering
	\includegraphics[width=0.9\textwidth]{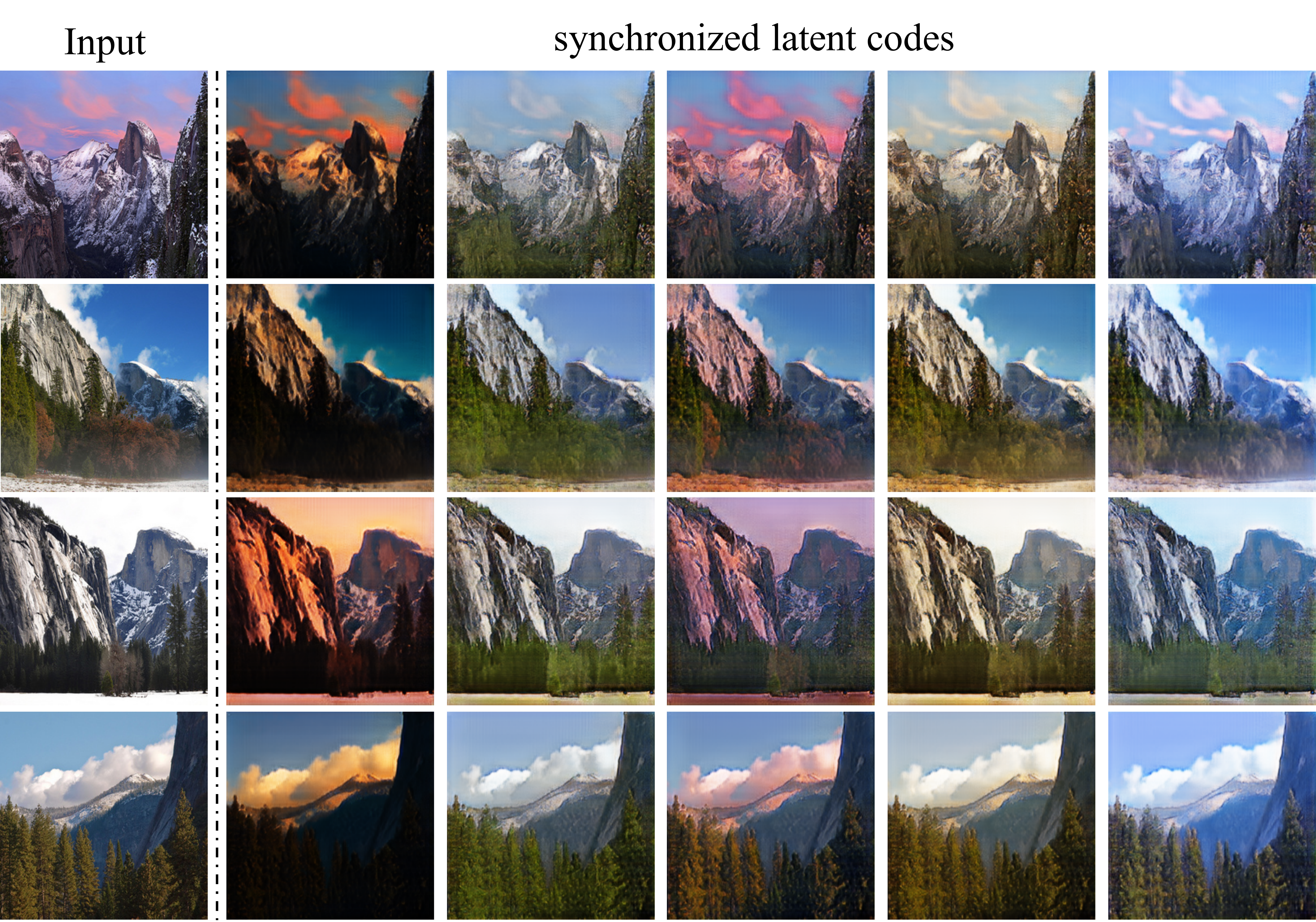}
	\caption{Disentangled results of our method on the Yosemite winter$\rightarrow$summer
		dataset under an unpaired setting. Each column's images are produced by the same latent code.}
	\label{fig:winter2summer}
\end{figure*}

\begin{figure*}[t!]
	\centering
	\includegraphics[width=0.86\textwidth]{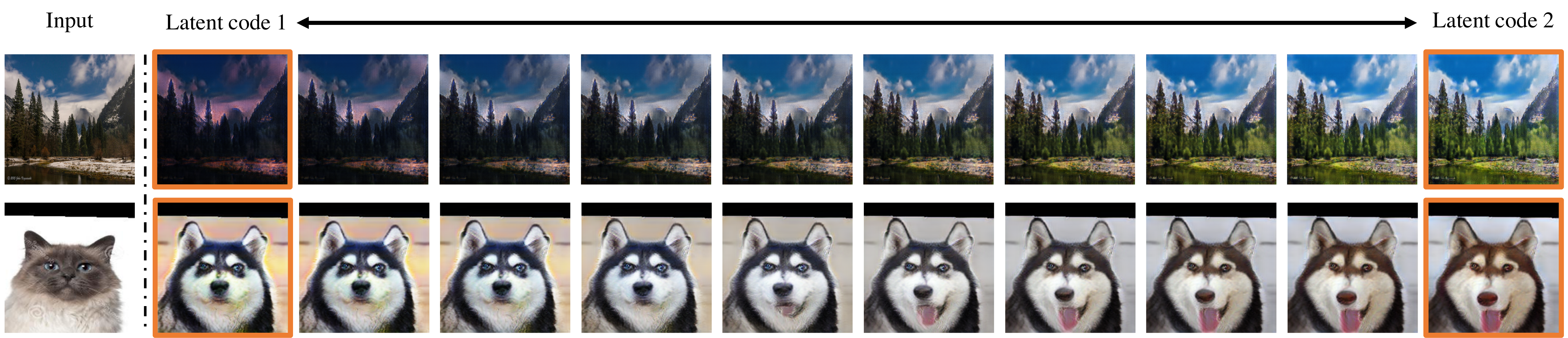}
	\caption{Translated samples generated from linear-interpolated vectors between two latent
		code vectors.}
	\label{fig:interpolation}
\end{figure*}

\subsection{Unpaired image-to-image translation}
Under an unpaired setting, we choose a one-sided unpaired I2IT method
GcGAN~\cite{fu2019geometry}
%\footnote{Note that theoretically, our method can be combined with
%DistanceGAN~\cite{benaim2017one} or CUT~\cite{park2020contrastive} as well.}
as the baseline (theoretically our method can also be combined with
DistanceGAN~\cite{benaim2017one} or CUT~\cite{park2020contrastive}), resulting in a much
simpler and fast-training framework compared with methods like
MUNIT~\cite{huang2018multimodal}, DRIT~\cite{lee2018diverse}, and
DRIT++~\cite{lee2020drit++} (see Figure \ref{fig:model}). For simplicity, we use a U-net generator
and a multi-scale discriminator as in BicycleGAN~\cite{zhu2017toward}. The U-net generator can also
help preserve the content as its skip-connections propagate the shape information directly from input
to output. We set the weight of geometry-consistency (GC) loss as 20. We find that GC loss is useful
for shape-invariant datasets like Yosemite winter$\rightarrow$summer, but may degrade the quality
for shape-variant datasets like cat$\rightarrow$dog. So we omit it when trained on
cat$\rightarrow$dog and find that the U-net generator architecture is sufficient for preserving the input
image's pose.

Figure \ref{fig:unpaired} and Table \ref{tab:unpaired_results} show different methods'
qualitative and quantitative results on Yosemite winter$\rightarrow$summer and cat$\rightarrow$dog
datasets. Our method surpasses the baseline method on diversity without loss of quality. While with
a much simpler network architecture and a smaller training budget than the state-of-the-art methods,
\eg, MUNIT and DRIT++, our method also beat them with a noticeable margin, verifying the robustness
of our method to improve diversity without loss of quality for I2IT.

Our method can also achieve disentanglement between the source domain content and
the target domain style for free. We produce synchronized results using the same code for each style
on the Yosemite winter$\rightarrow$summer dataset. In Figure \ref{fig:winter2summer}, it can be
seen that each latent code represents a uniform style regardless of the input image.
In addition, we perform linear interpolation between two given latent codes to show the generalization
of our method to capture the conditional distribution. In Figure \ref{fig:interpolation}, the interpolation
reflects smooth changes in semantic level, \eg, the daylight and the tongue patter changes gradually
with the  variations of the latent codes. More qualitative results can be found in supplementary
materials.

\section{Conclusions and discussions}
We have presented a novel method to encourage diversity for image-to-image translation by adopting
a deep neural estimator to estimate and maximize the mutual information between the latent code and
the generated image in cGANs. A simple network enhancement to cGANs is proved to be sufficient for
using our proposed loss. Our method can also disentangle target domain style from the source domain
content for free. Extensive qualitative and quantitative evaluations demonstrate the effectiveness of our
method to improve diversity without loss of quality.

Furthermore, we would like to discuss the wide application domains that may benefit from our method,
\eg, other conditional image synthesis tasks that desire diversity (like image inpainting, text-to-image
synthesis, and style transfer), disentangling latent factors in a conditional generative model setting, and
encouraging diversity in conditional image generation (like on ImageNet~\cite{deng2009imagenet}).

{\small
\bibliographystyle{ieee_fullname}
\bibliography{egbib}
}

\appendix
\cleardoublepage
\section{On the Latent Code Reconstruction Loss and the Variational Mutual Information Maximization}
\label{variational}
We show that the latent code reconstruction loss is closely related to the variational mutual information
maximization~\cite{barber2003algorithm,chen2016infogan}. The latent code reconstruction loss is
defined as follows:
\begin{equation}
\mathcal{L}_{R} = \mathbb{E}_{z}\left[||E(G(\cdot,z))-z)||_1\right]
\label{eq:reconstruction}
\end{equation}
where $z$ is the latent code, $\hat{b}=G(\cdot,z)$ is the generated image, and $G(\cdot, z)$ and
$E(\cdot)$ are the
generator's
and the encoder's  function respectively.

From \cite{chen2016infogan}, the variational lower bound on the mutual information between the latent
code $Z$ and the output
image $\hat{B}$ can be derived below:
\begin{equation}
\begin{split}
&\mathcal{I}(Z;\hat{B})=H(Z) - H(Z|\hat{B}) \\
=&\mathbb{E}_{\hat{b}\sim G(\cdot,z)}[\mathbb{E}_{z\sim p(z\vert\hat{b})}[\log
p(z\vert\hat{b})]]+ H(Z)\\
=&\mathbb{E}_{\hat{b}\sim G(\cdot,z)}[\underbrace{\KL[p(z|\hat{b})\Vert
	q(z\vert\hat{b})]}_{\mathbf{\ge 0}}
+
\mathbb{E}_{z\sim p(z\vert\hat{b})}[\log q(z\vert \hat{b})]] +
H(Z)\\
&\ge \underbrace{\mathbb{E}_{\hat{b}\sim G(\cdot,z),z\sim p(z)}[\log q(z\vert \hat{b})]}_{\text{latent
		code reconstruction loss}}+ H(Z)
\end{split}
\end{equation}
where $p(z|\hat{b})$ and $q(z|\hat{b})$ are the true
posterior distribution and the variational posterior distribution respectively. If we choose the variational
distribution $q(z|\hat{b})$ as a Laplace parameterized by the encoder, then we have
the the latent code reconstruction loss as shown in Equation \ref{eq:reconstruction}.

\section{On the Jensen-Shannon Divergence and Mutual Information}
\label{proof}
Although it has been pointed in \cite{hjelm2018learning}, for self-containment of the paper, we also
show the relation between the Jensen-Shannon Divergence (JSD) between the joint and the product of
marginals and the Mutual Information (MI). They are related by
Pointwise Mutual Information (PMI)
\begin{equation}
\text{PMI}(x;y)\equiv \log \frac{p(x,y)}{p(x)p(y)}=\log \frac{p(y|x)}{p(y)}
\end{equation}

\begin{equation}
\begin{split}
\text{MI} \equiv \mathcal{I}(X;Y) &=\KL\left[p(x,y)\Vert
p(x)p(y)\right]
\\
&= \mathbb{E}_{p(x,y)}\left[\log
\frac{p\left(x,y\right)}{p\left(x\right)p\left(y\right)}\right]  \\ &= \mathbb{E}_{p\left(x,y\right)}\left[
\text{PMI}\left(x;y\right) \right]
\end{split}
\end{equation}
Derivation from \cite{hjelm2018learning}, we have
\begin{equation}
\begin{split}
&\text{JSD}\left[p\left(x,y\right)\Vert p\left(x\right)p\left(y\right)\right] \propto \\
&\mathbb{E}_{
	p\left(x,y\right)}
\left[
\log \frac{p\left(y|x\right)}{p\left(y\right)}  -
\left(1+\frac{p(y)}{p(y|x)} \right)
\log \left(1+\frac{p\left(y|x\right)}{p\left(y\right)}\right) \right]
\end{split}
\label{eq:JSD}
\end{equation}
The quantity inside the expectation of Equation \ref{eq:JSD} is a concave, monotonically increasing
function of the ratio $\frac{p(y|x)}{p(y)}$, which is $e^{\text{PMI}(x,y)}$.  Therefore, positive
correlation exists between the MI and the JSD between the joint and the product of marginals.

\section{Network Architecture of the Statistics Network}
In this section, we show the network architecture of the statistics network. The latent code is replicated
to form a feature map with $|z|$ channels and the proper spatial size to concatenate with every
convolutional layer's input. The latent code is also concatenated with the last convolution layer's output
to feed into the first fully-connected layer.
The network architecture is shown in Table \ref{tab:netS_256}.

\section{Sampling Strategies}
We tried different sampling strategies for obtaining negative samples. A batchwise sampling strategy
which takes all the non-corresponding pairs within the batch as the negative samples would  easily run
out of GPU memory. While the sampling strategy which obtains
non-corresponding samples by resampling from one of the marginal distributions performs badly in our
experiments. Instead, following \cite{hjelm2018learning}, we exclude the positive samples from the
negative samples and resample negative samples with the same size as the positive samples by
generating another batch of images and sample non-corresponding latent codes for them. We find
such a sampling strategy is stable and memory-efficient in our experiments.

\section{Evaulation Details}
For the FID, similar to \cite{mao2019mode}, we randomly generate 50 images per input image in the
test set. We randomly choose 100 input images and their corresponding generated images to form
5,000 generated samples. We use the 5,000 generated samples and all samples in training set to
compute FID.

For the LPIPS distance, we follow the settings from \cite{zhu2017toward}. We use 100 input
images from the test set and sample 19 pairs of translated images for each input image. Then we
average over them.

For NDB and JSD, we follow the suggestion of \cite{richardson2018gans} and make sure there are at
least 10 training
samples for each cluster. And similar to \cite{mao2019mode}, we employ all the training samples for
clustering and choose $K = 20$ bins for the Facades dataset, and $K = 50$ for other datasets.

\section{More Qualitative Results under a Piared Setting}

\begin{figure*}[ht]
	\centering
	\includegraphics[width=\textwidth]{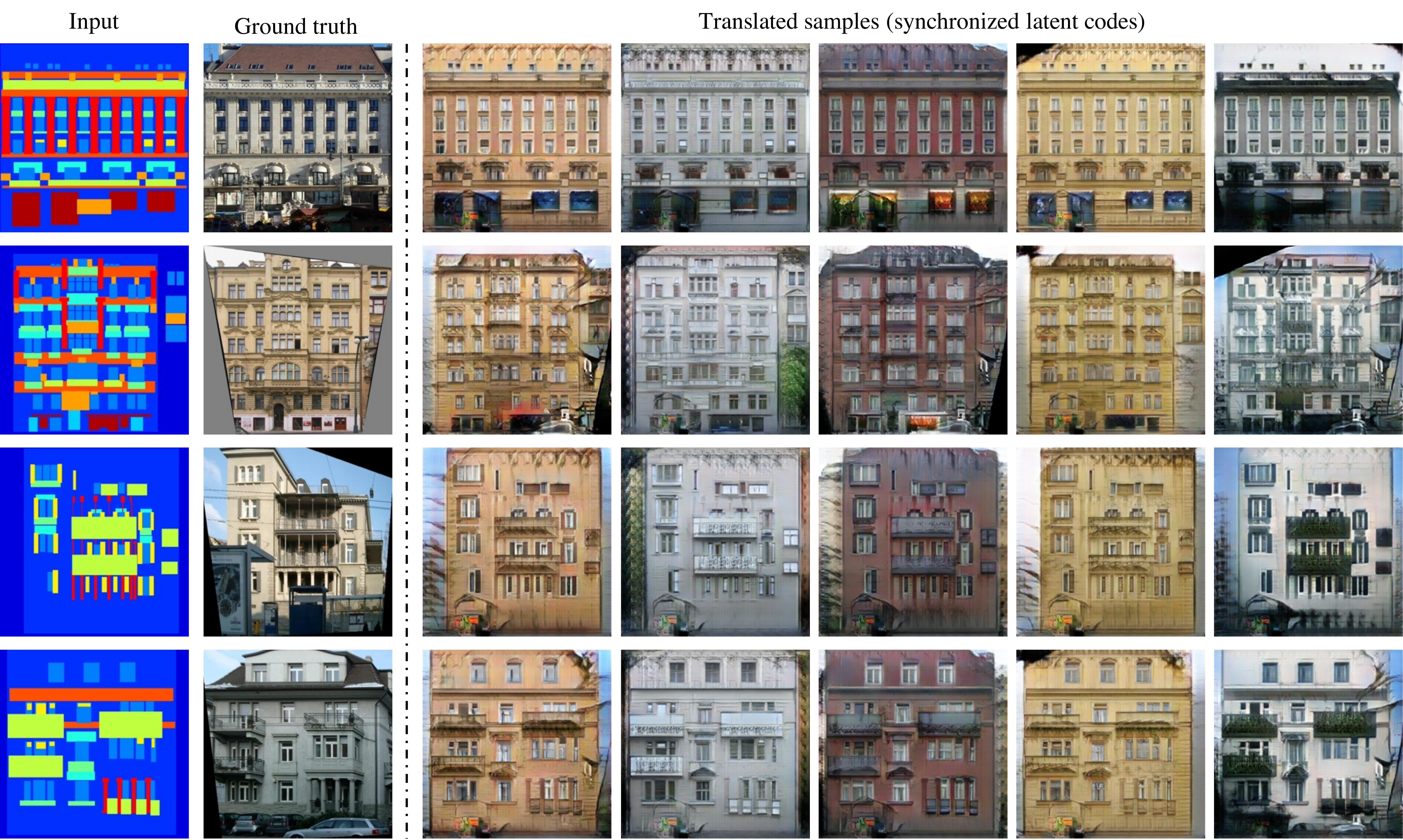}
	\caption{More qualitative results of our method on the Facades dataset under a paired setting.}
	\label{fig:facades_appendix}
\end{figure*}

\begin{figure*}[ht]
	\centering
	\includegraphics[width=\textwidth]{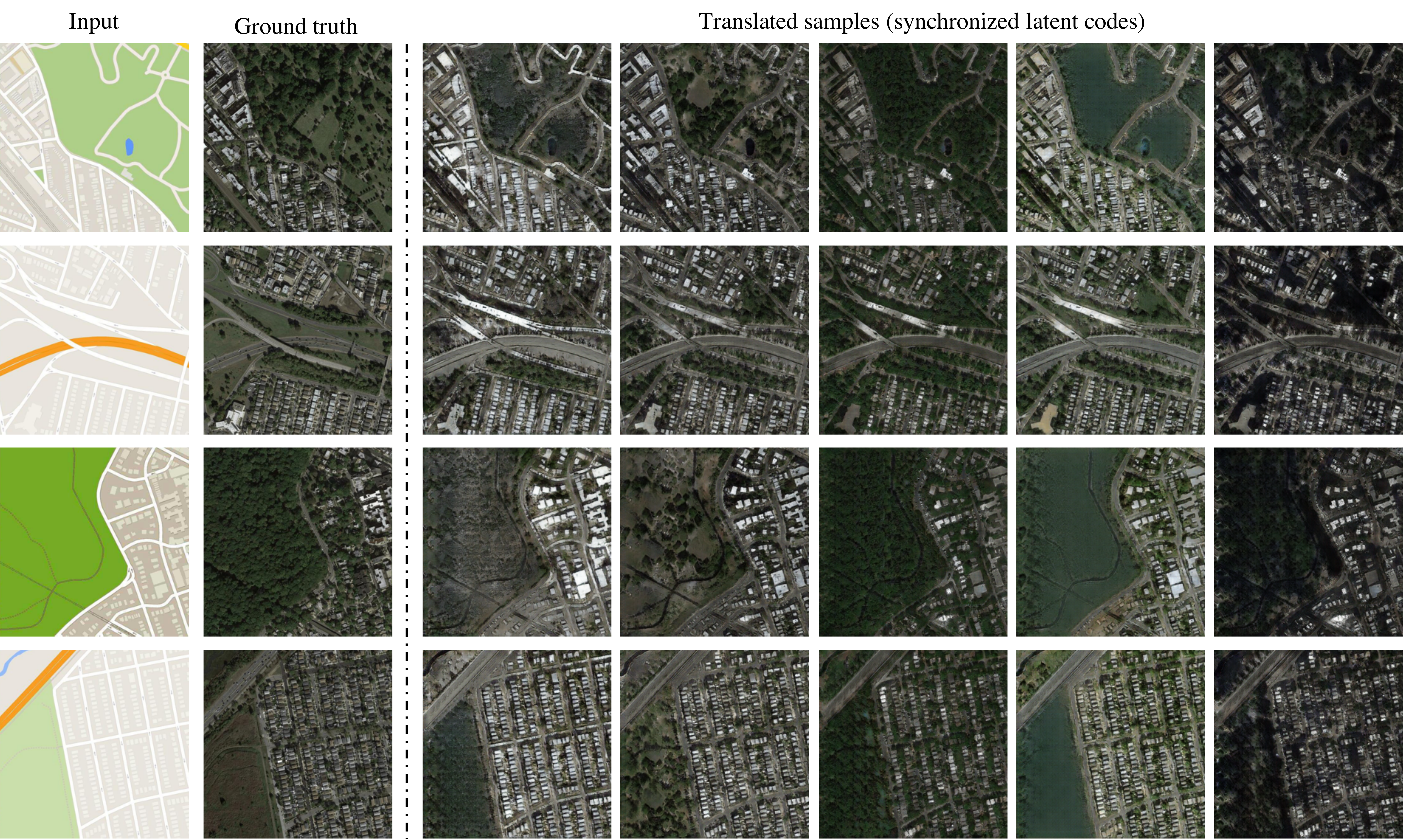}
	\caption{More qualitative results of our method on the Maps	dataset under a paired setting.}
	\label{fig:maps_appendix}
\end{figure*}

Here we display more qualitative results of our method on the Facades and the Maps datasets under a
paired image-to-image translation setting in Figure \ref{fig:facades_appendix} and
Figure \ref{fig:maps_appendix} respectively. All the translated samples are produced by
synchronized latent codes on different input images.

\begin{table*}[ht!]
	\centering
	\smallskip\begin{tabular}{c|c|c|c}
		\toprule
		\multicolumn{4}{c}{Statistics network}
		\\
		\hline
		\multicolumn{4}{c}{Input image $a\in \mathbb{R}^{256\times 256 \times 3}$ \& Latent code
			$z\in
			\mathbb{R}^8$}     \\
		\hline
		layer & input size & output size & non-linearity \\
		\hline
		4$\times$4 conv. stride 2.  & 3+8. & 32. & ELU.\\

		4$\times$4 conv. stride 2.  & 32+8. & 64. & ELU.\\

		4$\times$4 conv. stride 2.  & 64+8. & 128. & ELU.\\

		4$\times$4 conv. stride 2.  & 128+8. & 256. & ELU.\\

		4$\times$4 conv. stride 2.  & 256+8. & 512. & ELU.\\

		4$\times$4 conv. stride 2.  & 512+8. & 512. & ELU.\\
		\hline
		FC. & $4\times4\times512+8$. & 512. &  ELU.  \\
		FC. & 512. & 1. & none.\\
		\hline
	\end{tabular}
	\vspace{2mm}
	\caption{The statistics network's architecture.}\smallskip
	\label{tab:netS_256}
\end{table*}

\section{More Qualitative Results under an Unpaired Setting}
\begin{figure*}[ht]
	\centering
	\includegraphics[width=\textwidth]{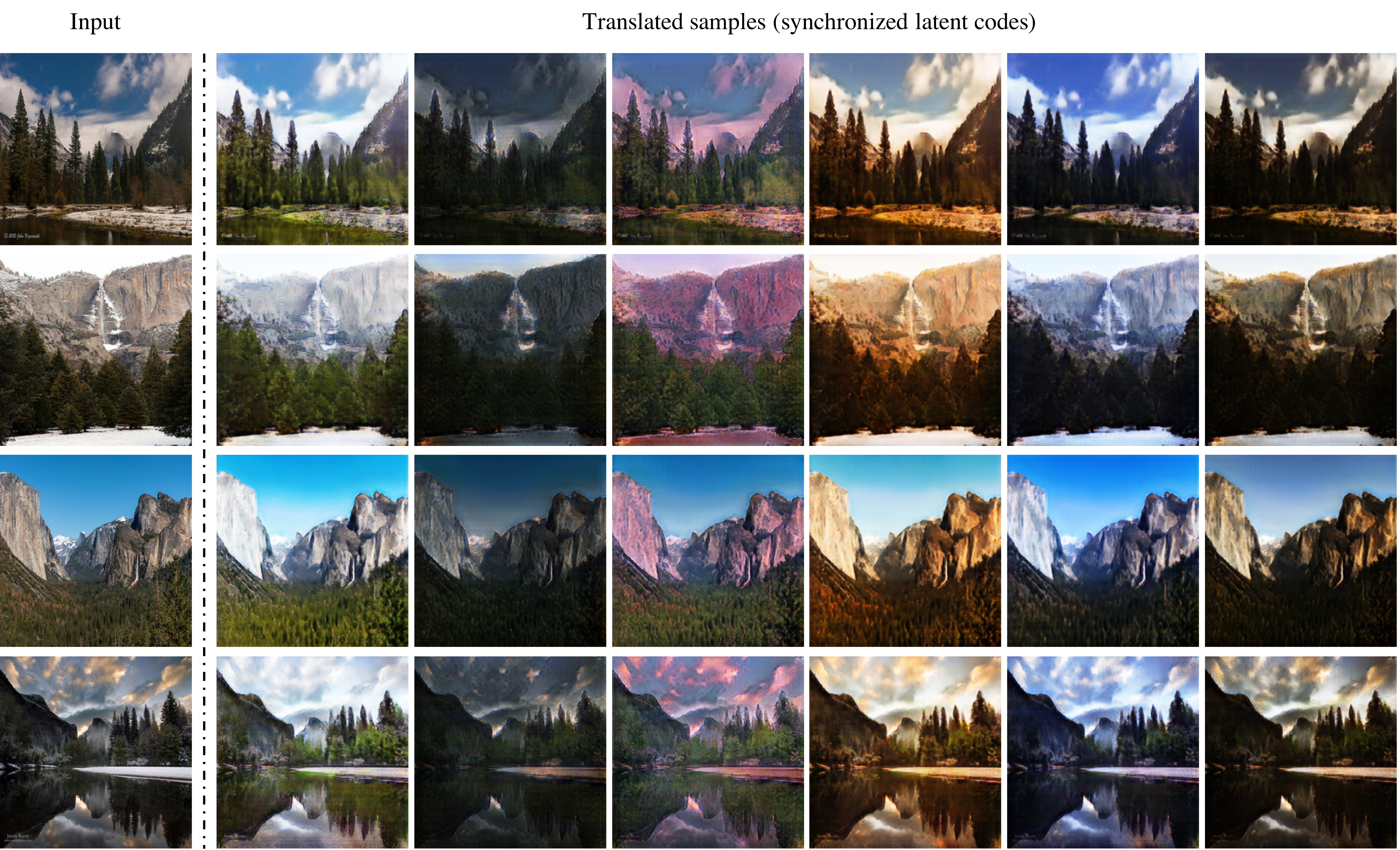}
	\caption{More qualitative results of our method on the Yosemite winter$\rightarrow$summer
		dataset
		under an
		unpaired setting.}
	\label{fig:winter2summer_appendix}
\end{figure*}

\begin{figure*}[ht]
	\centering
	\includegraphics[width=0.8\textwidth]{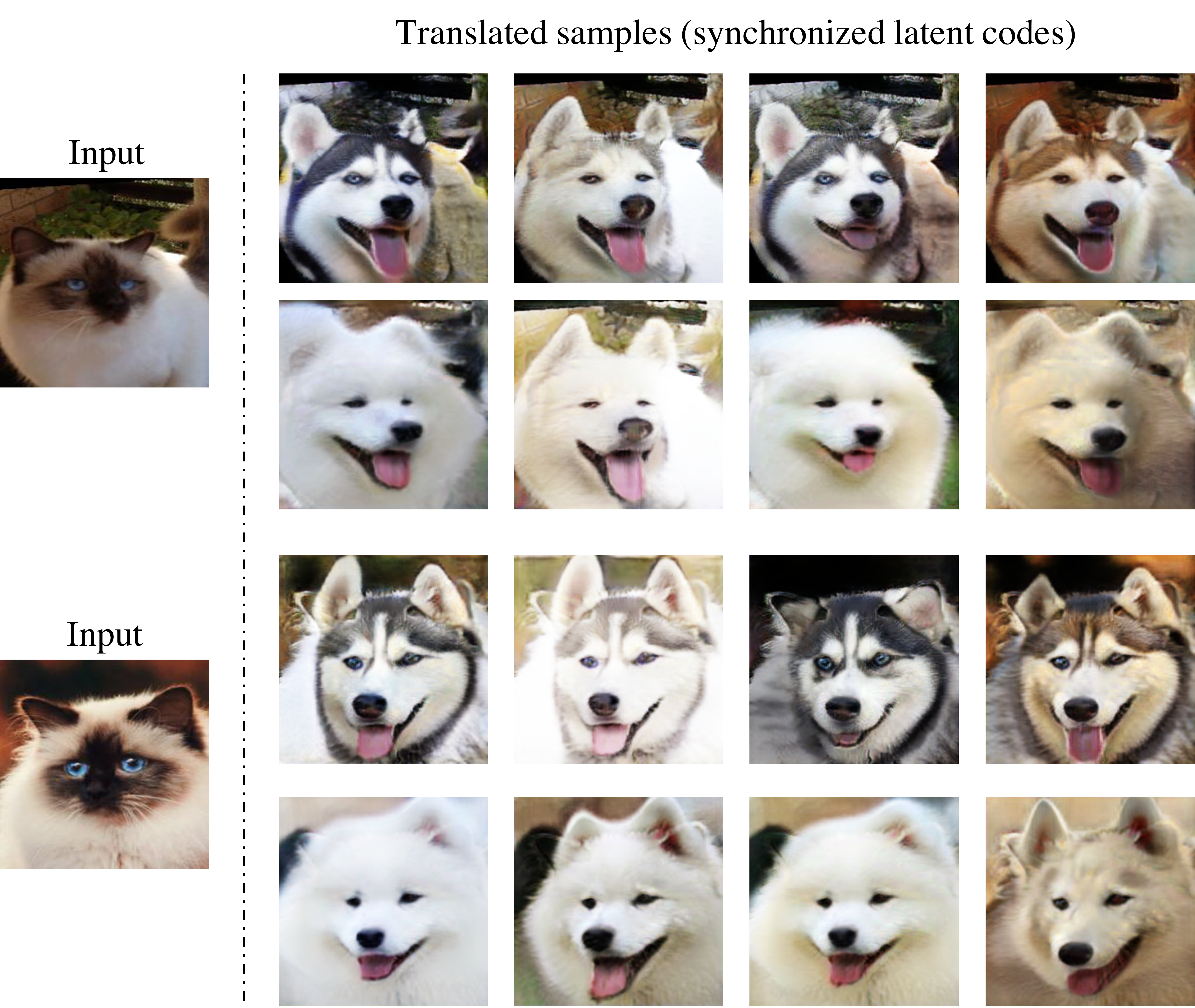}
	\caption{More qualitative results of our method on the cat$\rightarrow$dog dataset
		under an
		unpaired setting.}
	\label{fig:cat2dog_appendix}
\end{figure*}

More qualitative results of our method on the Yosemite winter$\rightarrow$summer and the
cat$\rightarrow$dog datasets under an unpaired setting are shown in Figure
\ref{fig:winter2summer_appendix} and Figure \ref{fig:cat2dog_appendix} respectively. All the
translated samples are produced by
synchronized latent codes on different input images.

\end{document}